\documentclass[hidelinks]{article}

\usepackage{arxiv}

\usepackage[utf8]{inputenc} %
\usepackage[T1]{fontenc}    %
\usepackage{hyperref}       %
\usepackage{url}            %
\usepackage{booktabs}       %
\usepackage{amsfonts}       %
\usepackage{nicefrac}       %
\usepackage{microtype}      %
\usepackage{lipsum}

\usepackage[superscript,biblabel]{cite}

\usepackage{tikz}
\usepackage{tikz-uml}
\usepackage{tikzscale}
\usetikzlibrary{arrows}
\usepackage{tikzscale}
\usepackage{textcomp}
\usepackage{soul,color}

\usepackage{tabularx}
\usepackage{booktabs}
\usepackage{colortbl} 
\usepackage{xcolor} 
\usepackage{xfrac}

\newcommand{\fmi}{\textsc{fmi}}
\newcommand{\api}{\textsc{api}}
\newcommand{\apis}{\textsc{api}s}
\newcommand{\mbd}{\textsc{mbd}}
\newcommand{\oop}{\textsc{oop}}
\newcommand{\aop}{\textsc{aop}}
\newcommand{\ros}{\textsc{ros}}
\newcommand{\yarp}{\textsc{yarp}}

\newcommand{\labview}{\textsc{Labview}}
\newcommand{\labviewr}{\textsc{Labview}\texttrademark}
\newcommand{\modelica}{\textsc{Modelica}}
\newcommand{\xcos}{\textsc{Xcos}}
\newcommand{\matlab}{\textsc{matlab}}

\newcommand{\simulink}{Simulink}
\newcommand{\simulinkr}{Simulink\textsuperscript{\scriptsize\textregistered}}
\newcommand{\simulinkcoder}{Simulink Coder}
\newcommand{\simulinkcoderr}{Simulink\textsuperscript{\scriptsize\textregistered} Coder\texttrademark}
\newcommand{\wbt}{\textsc{whole-body toolbox}}
\newcommand{\blockfactory}{\textsc{blockfactory}}
\newcommand{\cpp}{{C\nolinebreak[4]\hspace{-.05em}\raisebox{.4ex}{\tiny\bf ++}}}

\definecolor{background}{HTML}{ECECEC}
\definecolor{greyish}{HTML}{FEF3D2}
\definecolor{yellowish}{HTML}{FEF3D2}
\definecolor{yellow2}{HTML}{FBEF65}
\definecolor{pinkish}{HTML}{EDCFC0}
\definecolor{redish}{HTML}{F8AE98}
\definecolor{lblueish}{HTML}{C3D3D6}
\definecolor{mblueish}{HTML}{98B8BF}
\definecolor{dblueish}{HTML}{7398AA}
\definecolor{greenish}{HTML}{D5DFCF}
\definecolor{purpleish}{HTML}{C5AAAF}
\definecolor{dpurpleish}{HTML}{907C85}

\title{A Generic Synchronous Dataflow Architecture to Rapidly Prototype and Deploy Robot Controllers}

\author{
  Diego Ferigo\thanks{Dynamic Interaction Control, Istituto Italiano di Tecnologia, Genova, IT, 16163} ${}^{\ ,}$ \thanks{University of Manchester, Machine Learning and Optimisation, Manchester, UK, M13 9PL} \\
  \texttt{diego.ferigo@iit.it} \\
  \And
  Silvio Traversaro ${}^{*}$ \\
  \texttt{silvio.traversaro@iit.it} \\
  \And
  Francesco Romano ${}^{*}$\\
  \texttt{francesco.romano@iit.it} \\
  \And
  Daniele Pucci ${}^{*}$\\
  \texttt{daniele.pucci@iit.it}
}

\begin{document}
\maketitle

\begin{abstract}
The paper presents a software architecture to optimize the process of prototyping and deploying robot controllers that are synthesized using model-based design methodologies. The architecture is composed of a framework and a pipeline. Therefore, the contribution of the paper is twofold. First, we introduce an open-source actor-oriented framework that abstracts the common robotic uses of middlewares, optimizers, and simulators. Using this framework, we then present a pipeline that implements the model-based design methodology. The components of the proposed framework are generic, and they can be interfaced with any tool supporting model-based design. We demonstrate the effectiveness of the approach describing the application of the resulting synchronous dataflow architecture to the design of a balancing controller for the YARP-based humanoid robot iCub. This example exploits the interfacing with Simulink and Simulink Coder.
\end{abstract}


\section{Introduction}

In the past few decades, robotics has experienced a continuous shift from applications in constrained industrial environments to those involving autonomy, interaction, and collaboration with external agents. The adaptation of the robotic devices to new tasks often presents big challenges in both cost and time. Thus, the capability of prototyping a new controller and rapidly deploying it to the target robotic device is becoming more and more paramount. The canonical approach to develop a robotic controller can be summarized in two distinct phases\cite{schlegelDesignAbstractionProcesses2010}.
In the first phase, the robotic controller is synthesized, tuned, analyzed, and possibly tested in a simulated environment. Arbitrarily complex models of the controlled system are typically exploited. In the second phase, the controller is ported to the real device, tuned again and executed.
Each minor change to the controller requires iterating this entire process from start, and a lot of effort is spent to minimize manual operations.

Model-based design\cite{smithBestPracticesEstablishing2007} (\mbd) is a methodology that emerged to deal with the challenges introduced by the need to continuously improve complex systems. \mbd{} aims to simplify the development by providing a common environment shared by people of different disciplines involved in the the different design phases\cite{lennonModelbasedDesignMechatronic2008}. Later changes of the original design either due to early mistakes or requirements modifications are easier to propagate, therefore time and cost of the development can be reduced\cite{brugaliModelDrivenSoftwareEngineering2015}.
A characteristic of the model-based design is that the iterative process of the continuous improvement is performed with unified visual tools, typically based on dataflow programming languages and frameworks. The dataflow naming originates from the view of programs as directed graphs of computations, where the data flow between their components\cite{johnstonAdvancesDataflowProgramming2004}. Controllers are usually described and analyzed using block diagrams, and the view offered by dataflow programming, composed of nodes and edges, naturally translates to blocks and signals. The analogy between graphs and block diagrams makes \mbd{} particularly appropriate for controller design. A typical implementation of model-based design consists of the following stages:

\begin{enumerate}
    \item \emph{Plant modelling}: creation of a mathematical description of both the dynamics of the controlled system and its environment.
    \item \emph{Controller prototyping}: implementation of the operational aim of the controller acting on the plant model.
    \item \emph{System simulation}: assessment of the controller performance in a simulated environment containing the plant model.
    \item \emph{Controller deployment}: adaptation of the resulting controller to run online on the controlled system operating in the real environment.
\end{enumerate}

In the market there are many available open-source and commercial software that implement the specifications of \mbd{}. Most of them belong to a broader category of tools that fulfill the paradigm of actor-oriented programming\cite{aghaFoundationActorComputation1997} (\aop).
Contrarily to object-oriented programming (\oop), where data structures interact via procedure calls, in actor-oriented languages concurrent objects are the first-class citizens. These objects are also called \emph{actors} and they communicate with each other via predefined \emph{channels}. Actors have well-defined interfaces that abstract their internal state and define constraints on how they can interact with the outside. The interaction between actors is never direct as the channel mediates it. In this way, the actors are independent entities that are not directly connected with other actors.
The definition of \aop{} is very broad and many \emph{models of computation}\cite{leeActorOrientedDesignEmbedded2003} can be identified to categorize the nature of the interaction between actors and channels. Each of these models are characterized by constraints in their execution, typically in the form of internal computation, internal state update, external computation, and type of communication between actors. Among all the available models, \emph{synchronous dataflow}\cite{leeSynchronousDataFlow1987} is particularly suited for the design of robotic controllers. The computations performed by this model are triggered by the availability of new input data and the connections between actors are buffered. Examples of tools belonging to this domain are \simulinkr\cite{SimulinkSimulationModelBased}, \xcos\cite{XcosWwwScilab}, \modelica\textsuperscript{\scriptsize\textregistered}\cite{tillerIntroductionPhysicalModeling2012} and \labviewr\cite{LabVIEW2018National}.

The application of \mbd{} to robotic controllers development can narrow the gap between control engineers, used to approach systems with block diagrams, and software engineers, used to procedural and object-oriented programming. However, this approach is not exempt from the complications introduced by system integration, which often introduces time-consuming obstacles. Particularly for what concern robotics, actor-oriented programming languages by themselves are not the final solution. In fact, object-oriented programming still has a central role in the development of low-level algorithms. The aim of these actor-oriented languages is not substituting \oop, but complementing it. In fact \aop{} is more suitable to target the creation of applications that belong to higher abstraction layers, implementing a design principle compatible with the \emph{separation of concerns}\cite{mensSeparationConcernsSoftware2002,vanthienen5CbasedArchitecturalComposition2014}. It represents a valid choice to ease the interconnection of self-contained black-box functionalities, which represent the building blocks of any robotic controller.

In this work, we propose a software architecture composed of a framework inspired by \aop{} and a pipeline for its application to robot controllers design. The framework intends to reduce the effort spent on system integration while minimizing both code and functionality duplication. The pipeline implements all the \mbd{} stages and it aims to minimize the controllers lead time while automatizing as much as possible the prototyping and deployment processes. Rapid prototyping and continuous deployment are achieved interfacing the framework respectively with \simulink{} and \simulinkcoderr\cite{SimulinkCoder}. \simulink{} provides out-of-the-box a wide library of black-box functionality exposed as blocks and also allows to be extended and integrated with external algorithms. Its status of visual programming and debugging is very mature and well documented. \simulinkcoder{} provides the automatic code generation capability that aid the implementation of the deployment stage of \mbd{}, removing the need to port or adapt the controller to another domain before being executed in the target platform\cite{bruyninckxBRICSComponentModel2013}.

Despite our tools selection, the framework has been designed in such a way to simplify the integration with other existing actor-oriented frameworks. The logic of the presented black-box functions (developed in \oop) is independent of them. This design allows to effectively separate the two programming domains while exploiting the best features from both.

More specifically, the work presented in this paper is based on a previously introduced framework\cite{romanoWholeBodySoftwareAbstraction2017}. From the status described in that work, the underlying software architecture considerably changed, but a big effort was spent to maintain as much as possible the same user experience.
The whole-body interface layer proposed in the original work has been entirely removed, moving the responsibility of the robot abstraction to the middleware layer. Moreover, most of the improvements detailed in the same study have been adapted to the new architecture and implemented. Beyond a radical architectural advancement, the main extension presented in this work is the fulfillment of the automatic code generation support, fundamental to complete the implementation of model-based design.

The paper is structured as follows. First, we list tools and frameworks belonging to actor-oriented programming and implementing the model-based design pattern, and define a common terminology used throughout the paper. Then, we present the architecture of the proposed software framework and outline how we implemented \aop{} for robotic controllers design. Successively, we describe how the proposed framework can be exploited to obtain a pipeline that implements the typical stages of model-based design. We present a development cycle example for a balancing controller that targets a humanoid robot. We proceed by discussing the current limitations of this workflow and future improvements. Finally, we draw conclusions.

\section{Background}
\subsection{Related Software}

Model-based design is a methodology that covers many software layers. Following a top-down view, the conventional unified tools typical of \mbd{} usually share the following features:

\begin{itemize}
    \item Support to automatically generate real-time code from a model
    \item Present a graphical interface to visualize a model
    \item Serve an engine to initialize the directed graph of computation
    \item Provide a set of solvers to compute the output of each model element
    \item Include a library of default black-box functions
    \item Offer a set of \apis{} for interfacing with its engine
\end{itemize}

Providing a complete taxonomy of the existing tools and frameworks is not a trivial task. To simplify the analysis, we limit the overview to frameworks that use the synchronous dataflow model of computation, ignoring those that also support concurrent actors. Considering the scope of the current work, we separate the existing solutions in two categories: \emph{hybrid} and \emph{discrete-only}.

Hybrid tools are the most generic and complete, they typically allow performing both continuous and discrete simulations. Since they provide solvers for each of the two domains, hybrid tools can execute both offline simulations of continuous \textsc{ode} systems and their discrete equivalent which is compatible with real-time usage.

Discrete-only tools, instead, target only discrete-time systems, and their execution is limited to call an equivalent \texttt{step} function. Given their discrete nature, this second category is compatible by design with real-time usage.

Given these definitions, engines that belong to the hybrid group  are Drake\cite{tedrakeDrakePlanningControl2016}, OpenModelica\cite{OpenModelica}, and the commercial software \simulink, Dymola\textsuperscript{\scriptsize\textregistered}\cite{dassaultsystemesrDymola} and \labview\cite{LabVIEW2018National}. Excluding Drake, all the others are unified visual tools which fully enter into the model-based design framework. \simulink, in particular, is the engine that became the de-facto standard for model-based design. It implements all the \mbd{} stages providing great flexibility and very simple user experience. 

Other available engines are represented by those that emerged in the context of software engineering for robotics, all belonging to the category of the discrete-only tools. These type of engines are typically designed to support the development of software that runs in real-time on a robotic platform, and so they do not support simulation-specific features such as continuous time system modeling. Examples of this software are Stack of Task's Dynamic Graph\cite{mansardVersatileGeneralizedInverted}, Genom3\cite{malletGenoM3BuildingMiddlewareindependent2010}, OpenRTM\cite{andoSoftwarePlatformComponent2008} and Orocos\cite{bruyninckxRealtimeMotionControl2003}. A features comparison of the tools listed in this section is shown in Table~\ref{tab:comparison}.

\begin{table*}
    \centering
    \newcommand{\x}{\ensuremath{\times}}
    \newcolumntype{Y}{>{\centering\arraybackslash}X}
    \begin{tabularx}{1.0\textwidth}{lYYYYY}
        \toprule
        Software & Hybrid & Visual Tool & Code Generation & Real-time Native & Open Source \\
        \midrule \rowcolor{black!20}
        Drake         & \x &    &    & \x & \x \\
        OpenModelica  & \x & \x & \x &    & \x \\ \rowcolor{black!20}
        \simulink     & \x & \x & \x &    &    \\
        Dymola        & \x & \x & \x &    &    \\ \rowcolor{black!20}
        \labview      & \x & \x & \x &    &    \\
        \xcos{}       & \x & \x & \x &    & \x \\ \rowcolor{black!20}
        Dynamic Graph &    &    &    & \x & \x \\ 
        Genom3        &    &    &    & \x & \x \\ \rowcolor{black!20}
        OpenRTM       &    &    &    & \x & \x \\ 
        Orocos        &    &    &    & \x & \x \\ %
        \bottomrule
    \end{tabularx}
    
    \caption{Comparison of available synchronous dataflow frameworks. Hybrid frameworks support simulating both continuous \textsc{ODE} and discrete systems. Visual tools offer a graphical interface to create the computational graph with the actors. Executing standalone code typically involves the generation of a low-level description of the graph; visual frameworks might or might not support exporting the graph through code generation. Note that the listed non-visual tools are all C and C++ based, and their graph is already described in low-level languages. Furthermore, non-visual tools might or might not support real-time execution.}
    \label{tab:comparison}
\end{table*}

For what concerns the deployment stage of \mbd{}, we can identify few suitable frameworks that provide support of automatically generate code. The scope of this process is to convert a model prototyped as a directed graph to a low-level procedural representation. Nowadays, automatic code generation is a standard feature of the \matlab{} system. The \simulinkcoder{} toolbox allows generating optimized C and \cpp{} code from a \simulink{} model, and it provides support to customize the sources injecting custom code during the generation process. Other frameworks that are worth mentioning are most of the software suites based on the Modelica\cite{tillerIntroductionPhysicalModeling2012} language, which typically support generating low-level code from their models.

The Functional Mock-up Interface \cite{FunctionalMockupInterface} (\fmi), despite being outside the categorization described above, is still relevant to this overview. \fmi{} is a standardized interface widely used in industry for model-based development. It is a feature-rich and production-grade tool with a clear standard, constantly improving at each release but, as most of the tools listed in this section, it was not available when we started the development of our software stack. In any case, its adoption in its current form would not be possible due to the lack of the support of vector messages between actors, shortcoming that will be removed in the upcoming version of the standard.

Other interesting frameworks for controllers design which are related to the cited engines are the Robotic Toolbox\cite{corkeRoboticsVisionControl2017} and the Robotic System Toolbox\cite{RoboticsSystemToolbox}. The latter, particularly, is one of the few unified framework that fully implements \mbd{} specifically for robotic controllers. It is based on the \ros\cite{quigleyROSOpensourceRobot} middleware and it implements many of its features. However, the support of kinematics and dynamics has been added only recently and it lacks the possibility to be extended to interfacing with third-party robotic libraries.

\begin{figure}
    \centering
    \tikzset{every picture/.style={line width=0.75pt}} 

\begin{tikzpicture}[x=0.75pt,y=0.75pt,yscale=-1,xscale=1]

\draw  [color={rgb, 255:red, 0; green, 0; blue, 0 }  ,draw opacity=1 ][fill={rgb, 255:red, 236; green, 236; blue, 236 }  ,fill opacity=1 ] (260,110) -- (380,110) -- (380,180) -- (260,180) -- cycle ;
\draw  [fill={rgb, 255:red, 115; green, 152; blue, 170 }  ,fill opacity=1 ] (254.95,125.2) .. controls (254.95,122.33) and (257.2,120) .. (259.97,120) .. controls (262.75,120) and (265,122.33) .. (265,125.2) .. controls (265,128.07) and (262.75,130.4) .. (259.97,130.4) .. controls (257.2,130.4) and (254.95,128.07) .. (254.95,125.2) -- cycle ;
\draw  [fill={rgb, 255:red, 115; green, 152; blue, 170 }  ,fill opacity=1 ] (255,145.2) .. controls (255,142.33) and (257.25,140) .. (260.03,140) .. controls (262.8,140) and (265.05,142.33) .. (265.05,145.2) .. controls (265.05,148.07) and (262.8,150.4) .. (260.03,150.4) .. controls (257.25,150.4) and (255,148.07) .. (255,145.2) -- cycle ;
\draw  [fill={rgb, 255:red, 115; green, 152; blue, 170 }  ,fill opacity=1 ] (255,165.2) .. controls (255,162.33) and (257.25,160) .. (260.03,160) .. controls (262.8,160) and (265.05,162.33) .. (265.05,165.2) .. controls (265.05,168.07) and (262.8,170.4) .. (260.03,170.4) .. controls (257.25,170.4) and (255,168.07) .. (255,165.2) -- cycle ;
\draw  [fill={rgb, 255:red, 248; green, 174; blue, 152 }  ,fill opacity=1 ] (374.95,134.8) .. controls (374.95,131.93) and (377.2,129.6) .. (379.97,129.6) .. controls (382.75,129.6) and (385,131.93) .. (385,134.8) .. controls (385,137.67) and (382.75,140) .. (379.97,140) .. controls (377.2,140) and (374.95,137.67) .. (374.95,134.8) -- cycle ;
\draw  [fill={rgb, 255:red, 248; green, 174; blue, 152 }  ,fill opacity=1 ] (375,154.8) .. controls (375,151.93) and (377.25,149.6) .. (380.03,149.6) .. controls (382.8,149.6) and (385.05,151.93) .. (385.05,154.8) .. controls (385.05,157.67) and (382.8,160) .. (380.03,160) .. controls (377.25,160) and (375,157.67) .. (375,154.8) -- cycle ;
\draw    (220,125) -- (252.95,125.19) ;
\draw [shift={(254.95,125.2)}, rotate = 180.33] [fill={rgb, 255:red, 0; green, 0; blue, 0 }  ][line width=0.75]  [draw opacity=0] (8.93,-4.29) -- (0,0) -- (8.93,4.29) -- cycle    ;

\draw    (220,145) -- (253,145.19) ;
\draw [shift={(255,145.2)}, rotate = 180.33] [fill={rgb, 255:red, 0; green, 0; blue, 0 }  ][line width=0.75]  [draw opacity=0] (8.93,-4.29) -- (0,0) -- (8.93,4.29) -- cycle    ;

\draw    (220,165) -- (253,165.19) ;
\draw [shift={(255,165.2)}, rotate = 180.33] [fill={rgb, 255:red, 0; green, 0; blue, 0 }  ][line width=0.75]  [draw opacity=0] (8.93,-4.29) -- (0,0) -- (8.93,4.29) -- cycle    ;

\draw    (385,134.8) -- (418,134.99) ;
\draw [shift={(420,135)}, rotate = 180.33] [fill={rgb, 255:red, 0; green, 0; blue, 0 }  ][line width=0.75]  [draw opacity=0] (8.93,-4.29) -- (0,0) -- (8.93,4.29) -- cycle    ;

\draw    (385.05,154.8) -- (418.05,154.99) ;
\draw [shift={(420.05,155)}, rotate = 180.33] [fill={rgb, 255:red, 0; green, 0; blue, 0 }  ][line width=0.75]  [draw opacity=0] (8.93,-4.29) -- (0,0) -- (8.93,4.29) -- cycle    ;

\draw [color={rgb, 255:red, 109; green, 109; blue, 109 }  ,draw opacity=1 ]   (345,90) .. controls (369.89,95.25) and (315.91,125.03) .. (363.53,139.56) ;
\draw [shift={(365,140)}, rotate = 195.97] [color={rgb, 255:red, 109; green, 109; blue, 109 }  ,draw opacity=1 ][line width=0.75]    (10.93,-3.29) .. controls (6.95,-1.4) and (3.31,-0.3) .. (0,0) .. controls (3.31,0.3) and (6.95,1.4) .. (10.93,3.29)   ;

\draw [color={rgb, 255:red, 109; green, 109; blue, 109 }  ,draw opacity=1 ][line width=0.75]    (295,90) .. controls (270.39,95.25) and (323.73,124.7) .. (276.46,139.55) ;
\draw [shift={(275,140)}, rotate = 343.56] [color={rgb, 255:red, 109; green, 109; blue, 109 }  ,draw opacity=1 ][line width=0.75]    (10.93,-3.29) .. controls (6.95,-1.4) and (3.31,-0.3) .. (0,0) .. controls (3.31,0.3) and (6.95,1.4) .. (10.93,3.29)   ;

\draw  [fill={rgb, 255:red, 213; green, 223; blue, 207 }  ,fill opacity=1 ] (314.95,180.2) .. controls (314.95,177.33) and (317.2,175) .. (319.97,175) .. controls (322.75,175) and (325,177.33) .. (325,180.2) .. controls (325,183.07) and (322.75,185.4) .. (319.97,185.4) .. controls (317.2,185.4) and (314.95,183.07) .. (314.95,180.2) -- cycle ;
\draw    (320,205) -- (319.98,187.4) ;
\draw [shift={(319.97,185.4)}, rotate = 449.93] [color={rgb, 255:red, 0; green, 0; blue, 0 }  ][line width=0.75]    (10.93,-3.29) .. controls (6.95,-1.4) and (3.31,-0.3) .. (0,0) .. controls (3.31,0.3) and (6.95,1.4) .. (10.93,3.29)   ;

\draw (320,145) node  [align=left] {{\small Block}};
\draw (193.5,143) node  [align=left] {{\small Input}\\{\small Signals}};
\draw (448.5,143) node  [align=left] {{\small Output}\\{\small Signals}};
\draw (320,86) node  [align=left] {{\small Ports}};
\draw (320.5,214) node  [align=left] {{\small Parameters}};

\end{tikzpicture}
    \caption{Visualization of the implementation of an actor: the block.}
    \label{fig:block}
\end{figure}
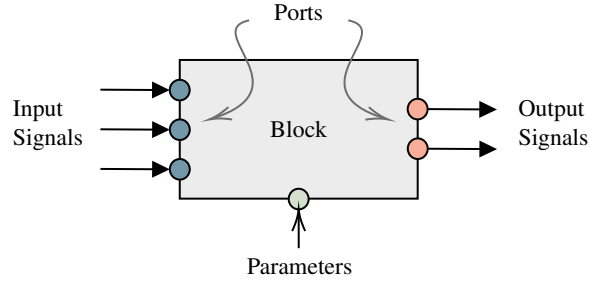

\begin{figure}
    \centering
    \resizebox{0.7\textwidth}{!}{
        \input{architecture.tikz}
    }
    
    \caption{Software architecture outline. \blockfactory{} provides the \texttt{Block} and \texttt{BlockInformation} \api{} along with the necessary tools to interface with the supported engines. The \blockfactory{} plugins contain the logic of the blocks, and they are loaded during runtime from the implementation of the engine \api. This figure illustrates well the abstraction of the components of the architecture. The engine is only aware of its own \api, which represent the entry-point that allow interfacing with it. The implementation of the engine \api{}, once blocks are loaded, can call their functionality only through the \texttt{Block} \api. Finally, the blocks contained in the plugins can communicate with the engine only through the abstraction provided by the \texttt{BlockInformation} \api.}
    \label{fig:architecture}
\end{figure}

\subsection{Terminology}

The majority of the hybrid and discrete-only software listed in the previous section share a software architecture composed of similar components. In view of the \aop{} architecture used in this work, we will make use of the following terminology, illustrated in Figure~\ref{fig:block}:

\begin{description}
    \item[Blocks] are elements that provide self-contained functionality. They wrap algorithms exposing a black-box interface composed of inputs, outputs, and parameters.
    \item[Ports] are virtual elements associated to block inputs and outputs. They store information to identify which kind of data is supported by the block (typically size and type).
    \item[Signals] are the elements that connect ports of different blocks. When two ports are connected, the share their data.
    \item[Engines] control the channel through which the blocks communicate. Engines typically create the computational graph and assign the blocks execution order. They also collect the block outputs and propagate them to the handled channel. They usually provide graphical tools to visualize blocks and help their interconnection by creating signals between them.
\end{description}

These terms naturally translate to the definitions of the \aop{} framework: blocks map to actors, signals map to channels, and ports represent the interface between actors and channels.

\section{Framework Software Architecture}

This section describes the software architecture of the proposed framework. Firstly, the factory pattern and the plugins concepts are introduced. Their combined usage has direct applicability within an \aop{} context. Secondly, we describe in details the two components that form the framework: \blockfactory\cite{dynamicinteractioncontrolBlockfactoryTinyFramework2019} and \wbt\cite{dynamicinteractioncontrolWholeBodyToolboxSimulink2019}. \blockfactory{} provides the support to actor-oriented programming and the interfacing with third-party frameworks. \wbt{} provides a plugin library containing the actors that expose the robotic stack used for controllers design: robotic middlewares, rigid-body dynamics libraries, and robotic simulators. An overview of the main classes of these two projects is shown in Figure~\ref{fig:uml}. In other terms, \wbt{} provides the algorithms, \blockfactory{} provides the back-end of the software infrastructure that abstracts blocks and engines. The solvers and the front-end are instead provided by the selected engine.

\begin{figure}
    \centering
    \resizebox{0.95\textwidth}{!}{
        \begin{tikzpicture}
\tikzumlset{font=\tiny, fill class=background}

\begin{umlpackage}[x=0,y=0,fill=dpurpleish!40]{blockfactory}
    \begin{umlpackage}[x=0,y=0,fill=purpleish]{core}
    
        \umlclass[x=0.6, y=0]{Signal}
        {
        }{
            + initializeBufferFromContiguous(BufferPtr, Length) : bool \\
            + initializeBufferFromNonContiguous(BufferPtr, Length) : bool \\
            + initializeBufferFromContiguousZeroCopy(BufferPtr, Length) : bool \\
            + isValid() : bool \\
            + getBuffer() : BufferPtr \\
            + get(Index) : Element \\
            + setBuffer(BufferPtr, Length) : bool \\
            + set(Index, Value) : bool
        }
        
        \umlclass[x=0, y=-8]{Port}
        {}{
            \umlstatic{+ DynamicSize : int}
        }
        
        \umlclass[x=-5, y=-2.5]{Parameter}
        {}{
            + isScalar() : bool \\
            + getScalarParameter() : Parameter \\
            + getVectorParameter() : VectorParameter \\
        }
        
        \umlclass[x=-5, y=-5]{Parameters}
        {}{
            + storeParameter(Parameter, ParameterMetadata) : bool \\
            + getParameter(Name, out: Parameter) : bool \\
            ...
        }
        
    
        \umlabstract[x=-4.3, y=-8.2, fill=yellowish]{Block}
        {
            \# m\_parameters : Parameters
        }{
            + getParameters(Parameters) : bool \\
            \umlvirt{+ parseParameters(BlockInformation) : bool} \\
            \umlvirt{+ configureSizeAndPorts(BlockInformation) : bool} \\
            \umlvirt{+ initialize(BlockInformation) : bool} \\
            \umlvirt{+ initializeInitialConditions(BlockInformation) : bool} \\
            \umlvirt{+ output(BlockInformation) : bool} \\
            \umlvirt{+ terminate(BlockInformation) : bool} \\
            ...
        }
        
        \umlabstract[x=0.7, y=-4, fill=greenish]{BlockInformation}
        {
        }{
            \umlvirt{+ addParameterMetadata(ParameterMetadata) : bool} \\
            \umlvirt{+ parseParameters(Parameters) : bool} \\
            \umlvirt{+ setPortsInfo(InputPortsInfo, OutputPortsInfo) : bool} \\
            \umlvirt{+ getInputPortInfo(PortIndex idx) : PortInfo} \\
            \umlvirt{+ getOutputPortInfo(PortIndex idx) : PortInfo} \\
            \umlvirt{+ getInputPortSignal(PortIndex) : InputSignal} \\
            \umlvirt{+ getOutputPortSignal(PortIndex) : OutputSignal} \\
            ...
        }
        
        \umlclass[x=-5.3,y=0]{ClassFactorySingleton}
        {}
        {
            + getClassFactory(ClassFactoryData) : bool \\
            + destroyFactory(ClassFactoryData) : bool 
            \\
            + extendSearchPath(Path) : void
        }
        
        \end{umlpackage}
        
        \begin{umlpackage}[x=7,y=-2, fill=redish]{simulink}
            \umlclass[width=30ex, fill=pinkish]{SimulinkBlockInformation}
            {}{}
        \end{umlpackage}
        
        \begin{umlpackage}[x=7,y=-6.5, fill=redish]{simulinkcoder}
            \umlclass[width=30ex, fill=pinkish]{SimulinkCoderBlockInformation}
            {}{
                + storeRTWParameters(Parameters) : bool \\
                + setInputPort(PortInfo, SignalAddress) : bool \\
                + setOutputPort(PortInfo, SignalAddress) : bool
            }
        \end{umlpackage}
        \end{umlpackage}
        
        \begin{umlpackage}[x=0.4,y=-12.8,fill=dblueish]{whole-body toolbox}
        
            \umlclass[x=-5,y=0,width=2.3cm,fill=lblueish]{ForwardDynamics}
            {}{
            }
            
            \umlclass[x=-2.5,y=0,width=2.3cm,fill=lblueish]{MassMatrix}
            {}{
            }
            
            \umlclass[x=0,y=0,width=2.3cm,fill=lblueish]{YarpRead}
            {}{
            }
            
            \umlclass[x=2.5,y=0,width=2.3cm,fill=lblueish]{QpOases}
            {}{
            }
            
            \umlclass[x=5,y=0,width=2.3cm,fill=lblueish]{SimulatorSynchronizer}
            {}{
            }
            
            \umlclass[x=7.5,y=0,width=2.3cm,fill=lblueish]{FooBlockName}
            {}{}
        \end{umlpackage}
        
        \umldep[geometry=-|,anchor1=30,anchor2=-140]{Block}{BlockInformation}
        \umlimpl[geometry=-|-]{SimulinkBlockInformation}{BlockInformation}
        \umlimpl[geometry=-|-]{SimulinkCoderBlockInformation}{BlockInformation}
        \umlimpl[geometry=|-|,anchor1=north]{QpOases}{Block}
        \umlimpl[geometry=|-|,anchor1=north,anchor2=south]{YarpRead}{Block}
        \umlimpl[geometry=|-|,anchor1=north,anchor2=south]{MassMatrix}{Block}
        \umlimpl[geometry=|-|,anchor1=north,anchor2=south]{SimulatorSynchronizer}{Block}
        \umlimpl[geometry=|-|,anchor1=north,anchor2=south]{FooBlockName}{Block}
        \umlimpl[geometry=|-|,anchor1=north,anchor2=south]{ForwardDynamics}{Block}
        \umldep[geometry=|-|, anchor1=40,anchor2=-60,arm1=1.3cm]{ForwardDynamics}{BlockInformation}
        \umldep[geometry=|-|, anchor1=40,anchor2=-60,arm1=1.3cm]{YarpRead}{BlockInformation}
        \umldep[geometry=|-|, anchor1=40,anchor2=-60,arm1=1.3cm]{QpOases}{BlockInformation}
        \umldep[geometry=|-|, anchor1=40,anchor2=-60,arm1=1.3cm]{MassMatrix}{BlockInformation}
        \umldep[geometry=|-|, anchor1=40,anchor2=-60,arm1=1.3cm]{SimulatorSynchronizer}{BlockInformation}
        \umldep[geometry=|-|, anchor1=40,anchor2=-60,arm1=1.3cm]{FooBlockName}{BlockInformation}
        \umldep[geometry=|-|,arm1=1.5cm]{Port}{BlockInformation}
        \umldep[geometry=|-,anchor1=30,anchor2=170]{Parameters}{BlockInformation}
        \umldep[geometry=|-|,arm2=3cm]{Signal}{BlockInformation}
        \umlcompo[geometry=|-|,arm2=0.5cm]{Block}{Parameters}
        \umlcompo[geometry=|-|,arm2=0.5cm]{Parameters}{Parameter}
        
    \end{tikzpicture}
    }
    \caption{\textsc{UML} diagram of the main classes part of the proposed architecture. The \blockfactory{} package is composed of three components: \texttt{core} contains the common classes and interfaces, \texttt{simulink} and \texttt{simulinkcoder} contain respectively the abstraction of the \simulink{} and \simulinkcoder{} engines.}
    \label{fig:uml}
\end{figure}

\subsection{Factory pattern and plugin libraries}

Third-party engines typically offer a set of \apis{} that can be used to integrate external software inside their framework. In order to detach effectively the block implementations from the third-party engine, the combination of the factory pattern and dynamically loaded plugins represents one of the canonical solution\cite{gammaDesignPatternsElements1995}. With the factory pattern, objects are created from a factory function without the need to specify their class. Typically a label or identifier is associated with this kind of objects, and only this information is required during their instantiation. Unfortunately, this is not enough to achieve the separation between engines and blocks, because the factory function only hides their allocation and the engine still needs to link against their implementation. This shortcoming can be overcome with plugin libraries dynamically loaded during runtime. In this case, the engine needs to have two information: the label associated with the implementation of the block and the name of the shared library that contains it. Once the plugin is dynamically loaded, the engine can instantiate block objects using a factory function without knowing anything about the class that implements them. Then, it can call their functionality through the common interface. The implementation of block classes is not constrained to any model of computation of \aop. In the most general form they can be asynchronous and concurrent.

The combined architecture of factory and plugins represents a natural implementation of actor-oriented programming. In fact, the limitation of the engine to access the functionality of the blocks through their exposed abstraction layer enforces one of the key characteristic of actors: the exposure of a well-defined interface.

For what concerns robotic controllers, the separation layer introduced by the plugin-based factory pattern provides a great help in system integration. In fact, since the plugin libraries containing the blocks are engine-agnostic, they can be loaded from each engine without the need to recompile them. This means that a controller prototyped with one engine can load the same library of the deployed controller. The code duplication is hence minimized and the robustness of the system is improved because the logic of the blocks is shared. Another benefit of this architecture to the system integration is about dependencies. The standalone plugins can link against any third-party library without the need to operate on the layer specific to the engine.

\subsection{BlockFactory}

The concepts defined by actor-oriented programming are implemented in a tool called \blockfactory. It allows creating blocks (the \emph{actors}) that exchange data between each other through the signals connected to their exposed ports, as illustrated in Figure~\ref{fig:block}. The entities of \aop{} are mapped to \cpp{} classes and interfaces, reported in Figure~\ref{fig:uml}. \blockfactory{} also implements the factory pattern and provides support to dynamically load during runtime plugins that contain block objects.

In order to obtain engine-agnostic blocks, the information exchanged between blocks and engines needs to be abstracted. For this scope, \blockfactory{} provides an abstraction layer called \texttt{BlockInformation} placed between blocks and engines. As shown in Figure~\ref{fig:architecture}, blocks can query information from the engine only through the \texttt{BlockInformation} interface, and engines can only call block functionalities through the \texttt{Block} interface.

The interfacing with third-part engines can be achieved in two steps. Firstly, their own \api{} or callbacks need to be implemented for loading during runtime the plugins containing the block logic. Secondly, in order to provide blocks the information from the engine they need, the \texttt{BlockInformation} interface needs to be implemented for the selected engine. In the current version of \blockfactory{}, we provide support of the \simulink{} and \simulinkcoder{} engines. In this case, the implementation of their \api{} corresponds in developing respectively a \textsc{c mex s}-function and a Target Language Compiler (\textsc{tlc}). \blockfactory{} provides these two files that are independent from the block implementation, and can load generic objects implementing the \texttt{Block} interface.

The actor-oriented applications that can be built with \blockfactory{} are universal, and not related by any means to robotic controllers. \blockfactory{} is engine-agnostic, and can be interfaced with engines specific to the target application. System integration is simplified since it contains only a small number of classes and it has no dependencies. Beyond the scope of the presented work, \blockfactory{} can find applicability in fields such as electrical drives, communication systems, power converters, etc. Generally, it can cover all use-cases that need exposing to the engines custom logic (either inlined in the block or wrapping external libraries) or interfacing with external devices while exposing only a simple and unified interface.

\subsection{Whole-Body Toolbox}

\wbt{} is a \cpp{} plugin library that exposes canonical algorithms and utilities commonly used to develop robotic controllers, such as rigid-body dynamics algorithms and communication capabilities with robotic devices mediated by middlewares. These functionalities are wrapped as block entities and they can be loaded independently by all the third-party engines supported by \blockfactory. In order to use the blocks in a \simulink{} model, the toolbox also provides a \simulink{} library that exposes the \cpp{} classes as visual blocks, which can be imported by drag-and-drop and configured through text boxes and drop-down menus.

For historical reasons the middleware we actively support is \yarp\cite{mettaYARPAnotherRobot2006}. Our main target platform is the iCub humanoid robot\cite{mettaICubHumanoidRobot2010}, even though all \yarp-compatible real and simulated robots are supported out-of-the-box. As an example, a previous work\cite{romanoWholeBodySoftwareAbstraction2017} showed a simulated whole-body controller running on both iCub and Walkman\cite{ferratiWalkManRobotSoftware2016} robots.

Historically \wbt{} was developed for whole-body control\cite{WholeBodyControlIEEE}, hence the name. In its last revisions, it became a generic robotic toolbox that can be used for any type of controller. The blocks implementing dynamics and kinematics algorithms are mainly based on iDynTree\cite{noriICubWholeBodyControl2015} and do not depend on any middleware. They can be used also with robots which are not \yarp-based, outsourcing, in this case, the interfacing with the target platform to third-party plugin libraries. The only requirement for using the provided algorithms is the availability of an \textsc{urdf}\footnote{http://www.ros.org/wiki/urdf} description of the robot to control.

A complete software stack for robotic controllers typically involves the interaction with a physic simulator. The robotic simulator we chose to support is Gazebo\cite{koenigDesignUseParadigms2004}.
The interaction between \simulink{} and Gazebo follows a co-simulation pattern, where the former is the master that issues forward step commands to the physic engine at each simulation step. The controller transparency between the real and the simulated robot is achieved by exposing the same network interface exploiting the abstraction layers provided by the \yarp{} middleware. In the case of the simulated robot, the implementation of these interfaces are provided by Gazebo Yarp Plugins\cite{mingohoffmanYarpBasedPlugins2014}.

The toolbox also provides generic utilities for robotic applications, such as discrete filters, cartesian trajectory controllers\cite{pattaciniExperimentalEvaluationNovel2010}, and quadratic programming solvers based on QpOASES\cite{ferreauQpOASESParametricActiveset2014}.

\begin{figure}
    \centering
    \resizebox{0.8\textwidth}{!}{
        \input{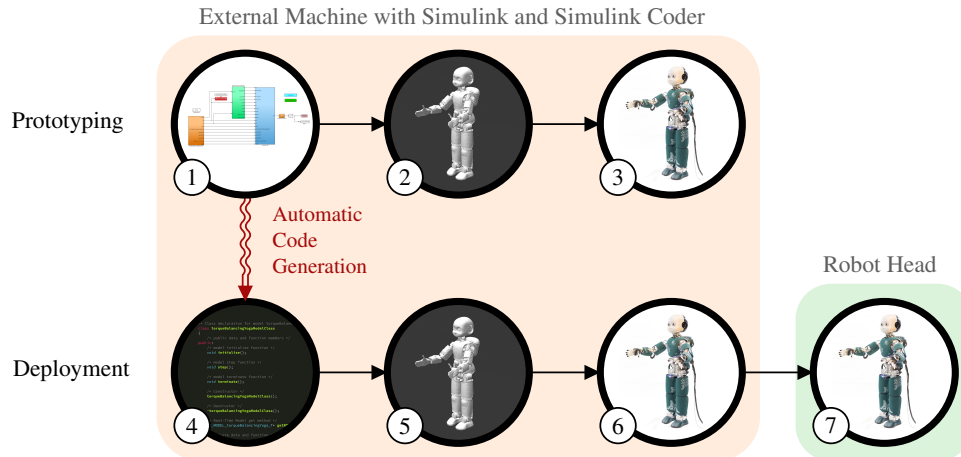}
    }
    \caption{Overview of the pipeline implementing model-based-design. The prototyping phase, in the first row, assumes the availability of a model of the robot. In step 1, a \simulink{} model of the controller is created. In step 2, the controller is tested in the Gazebo simulator using the robot model. In step 3, the same controller used in simulation is tested on the real robot, leveraging the robot transparency provided by exploiting the same \yarp{} interfaces. All the computations of this phase are executed from an external machine running \simulink. The communication with the real robot is achieved through the \yarp{} middleware. The second row illustrates the deploying phase. Exploiting \simulinkcoder, in step 4, C\hspace{-.05em}\raisebox{.4ex}{\tiny\bf++} code is automatically generated from the \simulink{} controller.
    Step 5 and 6 perform the same tests of the previous phase from the same external machine, respectively on the simulated and real robot. This time, though, the autogenerated controller is executed. Eventually, in step 7, the controller is deployed to the computer in the robot head and runs standalone.}
    \label{fig:pipeline}
\end{figure}

\section{The Pipeline}

In the previous section, we introduced the proposed framework, described its architecture, and discussed how its components interact with each other. In this section, we will describe the pipeline that implements \mbd{} from the point of view of the control engineer, detailing how the it is practically used and how the components of the framework relate to each step of the development.

The proposed pipeline implements all the four stages on which the model-based design pattern is derived. We will demonstrate a practical usage showing the steps to rapidly prototype and deploy a balancing controller\cite{navaExploitingFrictionTorque2018,pucciHighlyDynamicBalancing2016}, executed on the humanoid robot iCub\cite{mettaICubHumanoidRobot2010}. A simplified overview of the theory behind the controller is reported in the preceding work\cite{romanoWholeBodySoftwareAbstraction2017}. Since we managed to maintain the compatibility of the controllers designed with the previous architecture, the experimental results of that study that use \simulink{} correspond to the prototyping phase of this pipeline. As explained more in detail below, thanks to the abstraction between the controller and the robot provided by the \yarp{} interfaces, the pipeline includes few intermediate steps in addition to the stages defined by \mbd.

The first stage of \mbd{} is \emph{plant modeling}. For controllers applications, the plant is typically composed by the robot and the environment where it operates. In our case, the model of iCub is represented by an \textsc{urdf} file, which stores its kinematic and dynamic properties. The model of the robot is generated semi-automatically from its \textsc{cad} design, solution that allows obtaining a very detailed description of the robot. For what concern the environment, we use the default empty world provided by the physic engine running inside the simulator. The same applies to the interaction between the robot and the environment.

The implementation of the remaining stages of \mbd{} is illustrated in Figure~\ref{fig:pipeline}. The first row shows the prototyping phase of the pipeline and the second one shows the deploying phase. Referring to the figure, the depicted steps serve as follows:

\begin{enumerate}
\item This first step implements the \emph{controller prototyping} stage of \mbd. The controller is designed in \simulink{} using the default system blocks and the blocks provided by \wbt. In this case, when the user drops a block in the model, the \textsc{s}-function contained in \blockfactory{} loads the plugin library and, using the factory method, it allocates the object that implements its logic.
\item When the controller is ready, it can be executed on the simulated robot model. \wbt{} provides a block for interfacing with Gazebo, synchronizing it with the simulation loop running in \simulink. This step provides the means of the \emph{system simulation} stage.
\item In this additional third step, the control designer has the possibility to connect the controller, still running in \simulink{} from an external machine, to the real robot. Since the controller now needs to run in a real-time setting, the block used to interface with the simulator is substituted with a block that enforces the simulation loop to be synchronized with the real clock. Measurements and reference signals are gathered and streamed in real-time.
\item Reached this point, the controller is already functional on both the simulated and real robot. The last \emph{controller deployment} stage starts with step 4. Exploiting the capabilities of \simulinkcoder, the oriented graph visually created in \simulink{} is translated to an automatically generated \cpp{} class. In our software architecture, \simulinkcoder{} is handled as another engine (as reported in Figure~\ref{fig:uml}), and a different implementation of the \blockfactory{} interface that abstracts the engine is used. A very important detail of this process is that the logic implemented by the \wbt{} blocks is not inlined in the autogenerated class. In fact, analogously to the behavior of any engine supported by \blockfactory, the plugin-based factory pattern is used. This means that the autogenerated \cpp{} class loads the same plugin containing the logic of the robotic blocks that was used in the \simulink{} engine. Firstly, this helps to keep the behavior of the controllers running in different engines aligned. Secondly, assuming a constant controller graph, it simplifies the delivery of updates and fixes of the logic of the \wbt{} blocks. In fact, updated blocks can be deployed to the target platform by only distributing an updated plugin library, removing the need to regenerate the sources and rebuild the application. As last comment, it is worth noting that once the class has been generated and compiled, the presence of \simulink{} is no longer necessary.
\item This step corresponds to step 2. In this case, though, the automatically generated controller is executed on the simulated robot.
\item Similarly, this step corresponds to step 3 with the automatically generated controller.
\item The real deployment to the target platform is represented by this last step. Until now, the controller always ran from the external machine, communicating to the real robot through the network, exploiting the \yarp{} observer pattern\cite{gammaDesignPatternsElements1995}. The automatically generated class of the controller and the \blockfactory{} plugin are now compiled (or cross-compiled) for the on-board machine of the robot and, lastly, deployed. The comments about the choice of the plugin-based factory pattern of step 4 are even more central reached this last stage.
\end{enumerate}

In this example, the controlled robots -simulated and real- refer to the same kinematic structure. One may wonder which modifications are necessary in this new architecture in order to run the controller on a robot endowed with a different number of degrees-of-freedom. One of the new features of \wbt{} is the presence of a configuration block, where it is possible to specify runtime information such as the name of the \textsc{urdf} model, the names of the controlled joints, and the name of the robot used to set up the \yarp{} context. Excluding edge cases, this is enough to make controllers independent from the robot.

\section{Limitations and future work}

\subsection{Blockfactory}

The dataflow framework \blockfactory{} represents, as described in the previous sections, the abstraction layers between engines and black-box functions, supplied e.g. by plugins such as \wbt. This means that \blockfactory{} is responsible for exposing blocks in such a way that they can be properly configured by the solvers included in the engines. Currently, it only supports engines that provide discrete solvers with fixed-step. This is the only requirement for models that have to be executed on a real-time system. However, \blockfactory{} was born as a generic dataflow framework and, when the deployment is not the final target, it should provide compatibility with continuous solvers which typically need to operate on the derivatives of the block state.

At its current state, the block interface is modeled to be a stateless system. The engine can only trigger the evolution of a block state by calling its \texttt{output} method since they are akin to instantaneous functions. However, stateful blocks can be extremely convenient in some use cases. Indeed, \wbt{} already contains blocks that hold an internal state, but it is hidden inside the implementation. One of the consequences of the presence of this hidden state is that blocks that need to know the step size cannot gather it directly from the engine, and this information must be passed as a parameter. This behavior can be not very intuitive for the end user. Furthermore, is it more error prone since every time the user changes the step size, also the parameters of all blocks requiring it must be updated accordingly. This would not be necessary if the blocks would be modeled in such a way to expose their hidden state and rely on the engine features to address this shortcoming.

The Functional Mock-up Interface \cite{FunctionalMockupInterface} represents a common standard as an alternative to the provided interfaces. Instead of a complete substitution, though, being able to expose blocks in their counterparts called Functional Mock-up Units can be a valuable addition. This would open the interoperability with a plethora of tools that already support \fmi{}, improving the integration of the models designed with \blockfactory{} in complex co-simulation environments.

\subsection{Whole-Body Toolbox}

\wbt{} currently grounds the interfacing with robots on the \yarp{} middleware, and we are aware that there are not many existing \yarp{}-based robots. Despite implementing the \yarp{} interfaces for a new platform is not an insuperable task, it might limit the applicability of this pipeline. Going in this direction, a native implementation of the more common \ros{} middleware would enlarge the adoption of the proposed tools. A proof-of-concept of a \ros{} plugin implementing its publisher-subscriber pattern is already available\cite{ferigoBlockfactorydemorosProofofconceptPlugin2019}. On the same line, allowing to install the \wbt{} without its \yarp{} component would be another possible improvement. In fact, the majority of the blocks are middleware-agnostic, and they could be already used in systems without any middleware installed. For instance, many use-cases might benefit from the included algorithms for rigid body dynamics.

The current support of simulating a kinematic structure consists of a co-simulation setup between \simulink{} and Gazebo that communicate through \yarp{} messages thanks to the Gazebo Yarp Plugins. This entire system worked well for us in the past, however, its use is not as straightforward as it could be. In fact, in order to obtain a correct synchronization between the two simulators, all the components of this system should be started passing extra options. In order to simplify this process, it would be beneficial embedding the physic simulator inside a new block, treating it as a regular node of the graph. In this way, the synchronization could be greatly simplified taking advantage of the information available from being executed as part of the computational graph. This would also enable to execute headless simulations and allow to open the graphical user interface only if visual feedback is required.

A limitation of \wbt{} that might restrict its applicability to generic tasks is the lack of maturity of the robotic perception stack. The main scope of our applications are balancing and locomotion, therefore we always ignored perception and focused mainly on dynamics. Our controllers currently operate only on flat terrain, where perception is not required. However, creating new specific blocks to retrieve sensory data would be straightforward. An improved perception can then allow controllers to handle more structured scenarios, that can be already simulated in Gazebo inserting the robot model into a structured world.

In the long run, we would like to add the support of existing machine-learning frameworks in order to embed networks and function approximators into our robotic controllers. Furthermore, we are planning to introduce the possibility to export controllers with an interface that exposes a set of parameters which would allow applying reinforcement learning algorithms.

\subsection{Pipeline}

The description of the pipeline reported in the previous section offers a general overview of its functionalities. However, it hides few caveats which might not be straightforward. In step 2, obtaining a model that can be effectively actuated in Gazebo requires tuning its \textsc{pid} gains. Finding a proper configuration is not straightforward and many iterations are necessary. Furthermore, this process has to be repeated again in step 3, when the controller is executed on the real robot. Once the right gains have been properly found, they can be reused in steps 5 and 6. However, these low-level configurations are not strictly specific to this pipeline. In fact, they are related to the \yarp{} implementation of the robot and these parameters are meant to be abstracted by \yarp{} interfaces.

Similarly, it is interesting to analyze the factors that might differ between running the autogenerated controller from the external machine and from the on-board device of the robot. The communication between the controller and the robot ---typically consisting of sensor measurements and references--- are mediated in both cases by the transport layer handled by \yarp. In the first case, since the controller is running in an external machine, the exchange of data occurs through the network transport layer. This type of data transfer introduces overhead and delays that might affect the performance of the controller. Deploying the controller to the on-board machine provides a great opportunity to mitigate this problem. However, this is not exempt to side effects. In fact, controllers are very sensitive to time delays and dealing with them is yet an open problem in many applications. Assuming that the same gains can be applied might hold surprises. In our experience though, controllers did not need any tuning. In any case, moving the computation of fundamental tasks such as motion control as close as possible to the actuators offers a tremendous possibility to enhance system robustness. Furthermore, the deployed controller is an optimized version of the one executed in \simulink. If the rate of the controller is slower than the rate of the robot measurements and the actuation bandwidth, the gain of speed might allow increasing the controller frequency, which is typically related to better performances.

A current limitation of the autogeneration process is how controller parameters stored in \simulink{} are handled. With the current \blockfactory{} version, due to how the code is generated, accessing them from the code is not very intuitive. As a consequence, it is not yet possible to obtain an autogenerated controller that, without the need of regenerating the sources, can be used on different \yarp-based robots.

\section{Conclusions}

In this paper, we presented a rapid prototyping and deployment architecture for robotic controllers based on the principles of model-based design. The architecture is composed of a framework and a pipeline.

Developing and maintaining a controller in pure \cpp{} is typically extremely demanding, and even minor architectural changes might require a considerable effort. In light of the fast prototyping aims, developing controllers using visual tools and then automatically generate optimized \cpp{} code represents a great speedup.

As first component of the framework, we presented the actor-oriented tool \blockfactory{}. It abstracts generic algorithms and allows embedding them in generic applications modeled as directed graphs. The black-box functions that \blockfactory{} exposes are modeled as blocks with a predefined interface, and are stored in collections as shared libraries. These libraries can then be loaded from third-party software that implement the model-based design pattern. Among all the available possibilities, \blockfactory{} allows interfacing with \simulink{} and \simulinkcoder. However, it streamlines the extension to other frameworks by providing a second interface that abstracts the engines from the block implementations.

Different kind of robotic controllers are typically based on a limited set of elemental functionalities, and complex logic can be achieved by their composition. In this work, as second component of the framework, we presented \wbt, a collection of black-box functions for robotics representing the building elements of generic robotic controllers. This toolbox wraps a number of existing open-source projects belonging to the categories of robotic middlewares, rigid-body dynamic libraries, and mathematical optimization tools.

These two projects serve as the primary components of the proposed pipeline to rapidly prototype and deploy robotic controllers. In particular, the presented pipeline implements the rapid prototyping capability  ---idiomatic feature of model-based design--- by interfacing with the \simulink{} engine. The rapid deployment, instead, is achieved exploiting the automatic code generation support provided by \simulinkcoder. We explained step by step how the entire process works, detailing how the stages of model-based design have been implemented.

Ultimately, we explained the shortcomings of the current status of both the components of the presented framework and the resulting pipeline, and our plans to address them. The present condition of these projects is the outcome of many years of development, during which the architecture often changed and gave us the possibility to learn from our mistakes. Despite these continuous changes, a big effort has been spent to keep the experience of the control engineers that use this framework as consistent as possible. As attempted in the previous papers, we tried to be as critic as possible to our choices, being aware that the presented pipeline still has a big room of improvements.

To conclude, we would like to remark that the development of all the presented tools followed from their beginning an open-source and community-driven approach. From one hand, we could have never achieved the current development status and our results if we couldn't interface with existing open-source software such as middlewares, simulators, and libraries. We are grateful to the entire robotic community to provide and maintain them over time. From the other hand, collaborations with other research institutes --- mainly belonging to the community built around the iCub humanoid robot --- helped us to improve the robustness of the entire framework by using it within different contexts. A great contribution, as an example, regards the interfacing with \matlab. In fact, due to licensing limitation, we cannot test the pipeline thoroughly with many versions, and also the application of continuous integration pipelines presents many restrictions. A wider user-base with diverse setup helped us debugging problems we would probably have never encountered.

\section*{Acknowledgments}

This project has received funding from the European Union’s Horizon 2020 research and innovation programme under grant agreement No. 731540 (An.Dy).\newline
The content of this publication is the sole responsibility of the authors. The European Commission or its services cannot be held responsible for any use that may be made of the information it contains.

\bibliographystyle{vancouver}
\bibliography{references}  %

\end{document}